%% file: acl_latex.tex
\pdfoutput=1

\documentclass[11pt]{article}

\usepackage[]{acl}

\usepackage{times}
\usepackage{latexsym}
\usepackage{float, lscape} 
\usepackage{cleveref}
\usepackage[T1]{fontenc}
\usepackage{placeins}
\usepackage[utf8]{inputenc}
\usepackage{hyperref}
\usepackage{tabularx}
\usepackage{microtype}
\usepackage{booktabs}

\usepackage{graphicx}
\usepackage{caption}
\usepackage{subcaption}
\usepackage{xcolor}
\usepackage{multirow}
\usepackage{todonotes}
\usepackage{booktabs}
\usepackage{threeparttable}
\newcommand{\ggrev}[2]{\textcolor{black}{#2}}

\title{GECTurk: Grammatical Error Correction and Detection Dataset for Turkish}

\usepackage{algorithmicx}
\usepackage[ruled]{algorithm}
\usepackage[noend]{algpseudocode}

\author{Atakan Kara,~
Farrin Marouf Safian,~
Andrew Bond,~
Gözde Gül Şahin
\\[.3em]
Computer Engineering Department\\
Koç University, Istanbul, Turkey\\
{\url{https://gglab-ku.github.io/}}\\
}

\begin{document}
\maketitle
\begin{abstract}

Grammatical Error Detection and Correction (GEC) tools have proven useful for native speakers and second language learners. Developing such tools requires a large amount of parallel, annotated data, which is unavailable for most languages. Synthetic data generation is a common practice to overcome the scarcity of such data. However, it is not straightforward for morphologically rich languages like Turkish due to complex writing rules that require phonological, morphological, and syntactic information. In this work, we present a flexible and extensible synthetic data generation pipeline for Turkish covering more than 20 expert-curated grammar and spelling rules (a.k.a., writing rules) implemented through complex transformation functions. Using this pipeline, we derive 130,000 high-quality parallel sentences from professionally edited articles. Additionally, we create a more realistic test set by manually annotating a set of movie reviews. We implement three baselines formulating the task as i) neural machine translation, ii) sequence tagging, and iii) prefix tuning with a pretrained decoder-only model, achieving strong results. Furthermore, we perform exhaustive experiments on out-of-domain datasets to gain insights on the transferability and robustness of the proposed approaches. Our results suggest that our corpus, GECTurk, is high-quality and allows knowledge transfer for the out-of-domain setting. To encourage further research on Turkish GEC, we release our datasets, baseline models, and the synthetic data generation pipeline at \url{https://github.com/GGLAB-KU/gecturk}.
\end{abstract}

\section{Introduction}
\input{sections/intro}

\begin{figure*}
\centering
\includegraphics[width=1\linewidth]{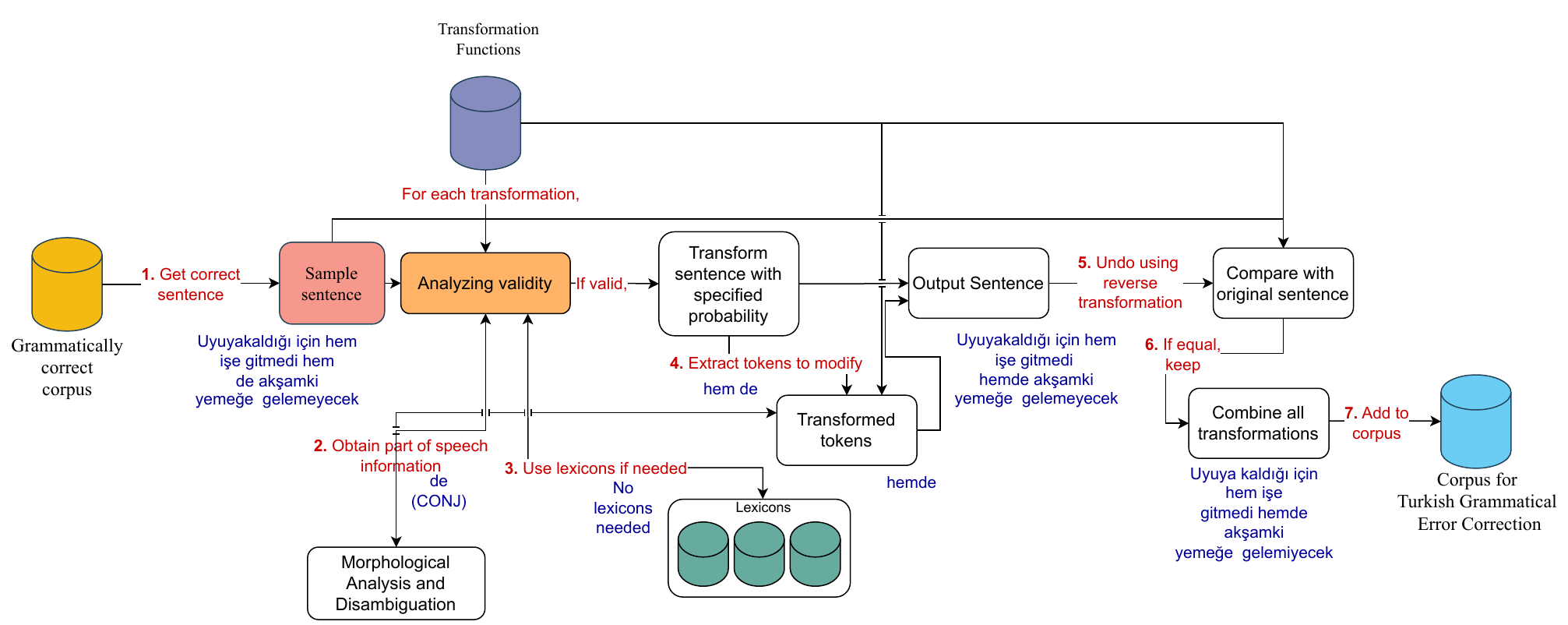}
\caption{Data generation pipeline. 1) First, a correct sentence is obtained from the grammatically correct corpus. 2) Then, morphological analysis is performed. 3) The validity of the sentence for the transformation function is checked. If the sentence is eligible, the transformation is applied with some probability $p$.  4)First, selecting tokens to modify, 5) then, checking if the reverse transformations can recover the original tokens. 6) If original cannot be recovered, the sentence is removed. If can be recovered 7) the transformed sentence is added to the corpus.}
\label{fig:pipeline}
\end{figure*}

\section{Related Work}
\input{sections/relatedWork}

\section{Synthetic Data Generation}
\input{sections/methodology}

\input{tables/trans_rules}

\section{Tasks and Models}
\input{sections/models.tex}

\section{Experimental Setup}
\label{sec:setup}
\input{sections/setup}

\begin{table*}[!htbp]
    \centering
    \scalebox{0.75}{
    \resizebox{\textwidth}{!}{
    \begin{tabular}{ccccccc}
        \toprule
        \multicolumn{7}{c}{\textbf{GECTurk}}\\
        & \multicolumn{3}{c}{\textbf{Detection}} & \multicolumn{3}{c}{\textbf{Correction}}\\
        \cmidrule(lr){2-4} \cmidrule(lr){5-7}
        
        & \textbf{P} & \textbf{R} & \textbf{$F_{1}$} & \textbf{P} & \textbf{R} & $\mathbf{F_{0.5}}$\\
        \midrule
        
        NMT     & -           & -           & -           & 0.50 $\pm$ 0.01 & 0.84 $\pm$ 0.01 & 0.55 $\pm$ 0.01 \\
        SeqTag  & 0.90 $\pm$ 0.003 & 0.90 $\pm$ 0.013 & 0.90 $\pm$ 0.006 & 0.98 $\pm$ 0.001 & 0.98 $\pm$ 0.001 & 0.98 $\pm$ 0.001 \\
        mGPT    & 0.52 $\pm$ 0.02 & 0.38 $\pm$ 0.004 & 0.41 $\pm$ 0.01 & 0.95 $\pm$ 0.01 & 0.92 $\pm$ 0.03 & 0.94 $\pm$ 0.01 \\
        
        \midrule
        \multicolumn{7}{c}{\textbf{Curated Test Data}}\\
        \midrule
        NMT     & -           & -           & -           & 0.31 & 0.62 & 0.35 \\
        SeqTag  & 0.94 & 0.87 & 0.89 & 0.85 & 0.80 & 0.84 \\
        mGPT    & 0.73 & 0.52 & 0.59 & 0.75 & 0.61 & 0.72 \\
        
        \bottomrule
    \end{tabular}
    }
    }
    \caption{Detection and Correction results of the baselines on GECTurk (in-domain) and curated test dataset (out-of-domain).}
    \label{table:main_results}
\end{table*}

\section{Experiments and Results}
\input{sections/results}

\section{Conclusion and Future Work}
\input{sections/conclusion}

\section*{Acknowledgements}
    This work has been supported by the Scientific and Technological Research Council of Türkiye~(TÜBİTAK) as part of the project ``Automatic Learning of Procedural Language from Natural Language Instructions for Intelligent Assistance'' with the number 121C132. We also gratefully acknowledge KUIS AI Lab for providing computational support. We thank our anonymous reviewers and the members of GGLab who helped us improve this paper. 
    
\section*{Limitations}
\input{sections/limitations.tex}

\section*{Ethical Considerations}
\input{sections/ethics}

\bibliography{anthology,custom}
\bibliographystyle{acl_natbib}

\clearpage
\appendix
\section{Model Details}
\label{app:models}

\subsection{NMT Baseline}
For tokenization, we used BerTurk-cased \cite{BERTurk} tokenizer, passed to the NMT model. The transformer model has 6 encoders with embedding size 512, 6 decoder layers, and 8 heads. A dropout of 0.1 is used directly after the positional embeddings. For training, an Adam \cite{Kingma2014AdamAM} optimizer with $\beta_{1} = 0.9$, $\beta_{2}=0.98$, and $\varepsilon=1e-9$, and a learning rate of $1e-4$ is used. We used batch size of 32, and trained the model for 100 epochs on a single V100.  We use a standard cross-entropy loss during training, as follows:
\begin{equation}
    \mathcal{L}_{GEC}(\hat{y}, y) = -\sum_{n=1}^{N}\sum_{c=1}^{V} log \frac{exp(\hat{y}_{n, c})y_{n, c}}{\sum_{i=1}^{V} exp(\hat{y}_{n, i})} 
\label{eq1}
\end{equation}
Here, $N$ is the batch size, $V$ is the number of error classes, $x$ is the model output, and $y$ is the target. For the data size experiments, we used the same architecture but with slightly different hyperparameters. For both the 75\% and 100\% experiments, the model was trained for 100 epochs. For the 50\% experiment, we only trained for 50 epochs. When training 10\% and 25\%, the Adam optimizer is used with the same $\beta$ values, a learning rate of $5e-4$, and a weight decay of $1e-4$, for 100 epochs. The zero-shot testing on the BOUN \cite{detecting_clitics} dataset is tokenized with the same tokenizer, and the best pre-trained model from GECTurk is used for evaluation.

\subsection{Sequence Tagger}
    For training, we used the AdamW~\cite{adamw} optimizer for $3$ epochs, using batch size $16$, learning rate $2e-5$, weight decay $0.01$, $\beta_{1}=0.9$, and $\beta_{2}=0.999$. 

\subsection{Prefix Tuning}
    We used the standard mGPT tokenizer and the OpenPrompt prefix tuning template. All experiments use 5 soft tokens at the beginning. Teacher forcing is used during training, and both the correction and detection tasks are formulated as a sequence generation problem. Following the settings from \citet{sahinMorph22}, we don't use weight decay for the bias and LayerNorm weights. The AdamW optimizer is used, with an initial learning rate of $5e-5$, linearly decaying to $0$ over the entire training. We clip the norm of the gradient at $1.0$. Due to the computational requirements of mGPT, we only train on GECTurk for a single epoch on all experiments. However, on the smaller BOUN dataset, we train for 5 epochs. For inference, we also follow the hyperparameters from \citet{sahinMorph22}, using a temperature of $1.0$, top p of $0.9$, no repetition penalty, and a beam search of 5 beams. For all experiments, a batch size of $3$ was used. The max sequence length, including soft tokens, is set to 512. 

\section{GECTurk Error Frequencies}
\label{sec:errorfreq}
\begin{figure}
        \centering
        \includegraphics[width=1\linewidth]{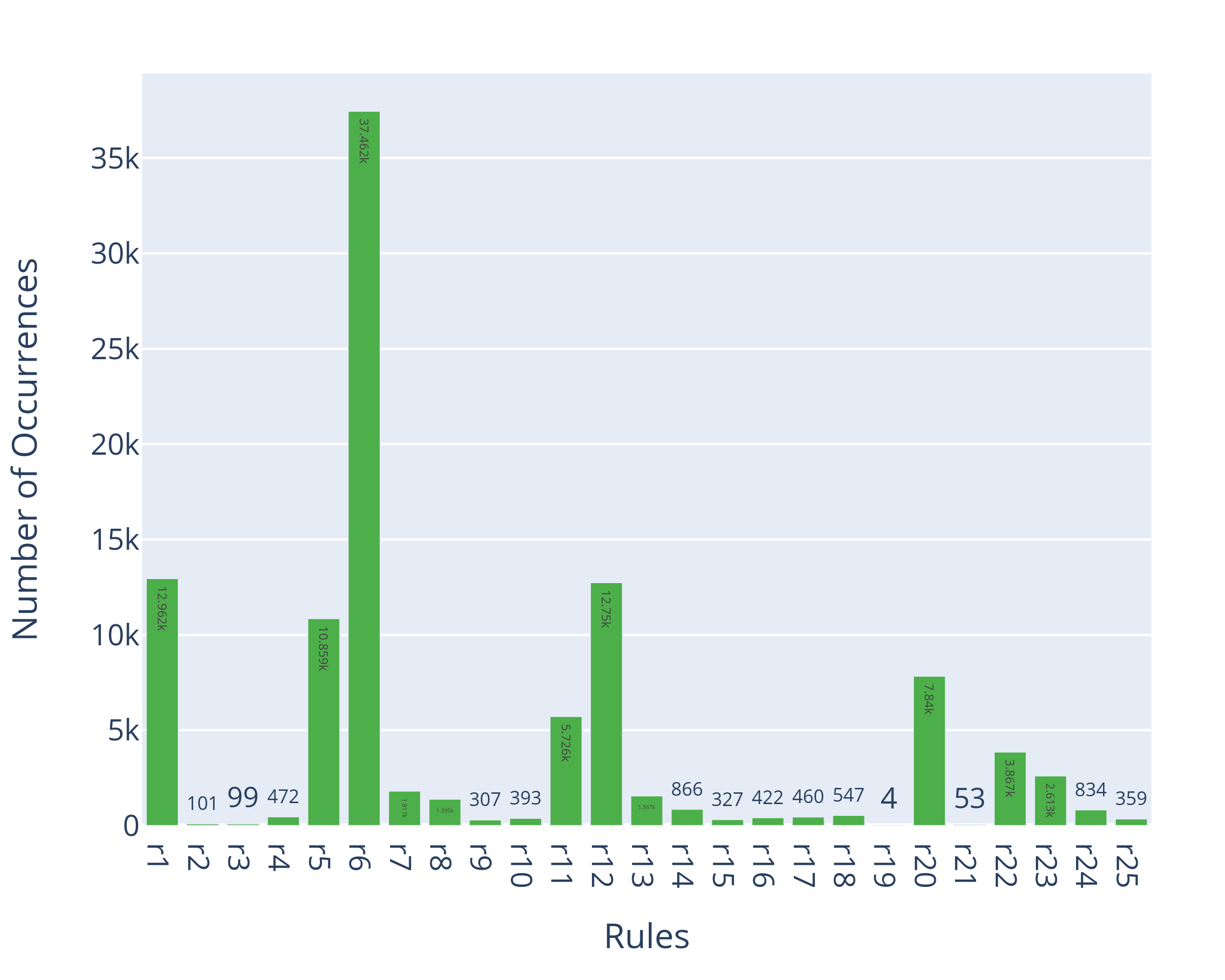}
        \caption{Number of sentences with each writing rule type.}
        \label{fig:dataset-sizes}
    \end{figure}

    Fig.~\ref{fig:dataset-sizes} shows the frequencies of each error type in GECTurk dataset.

\section{Effect of Dataset Size}
\label{appendix:dataset_size}
\begin{figure}
        \centering
        \includegraphics[width=1\linewidth]{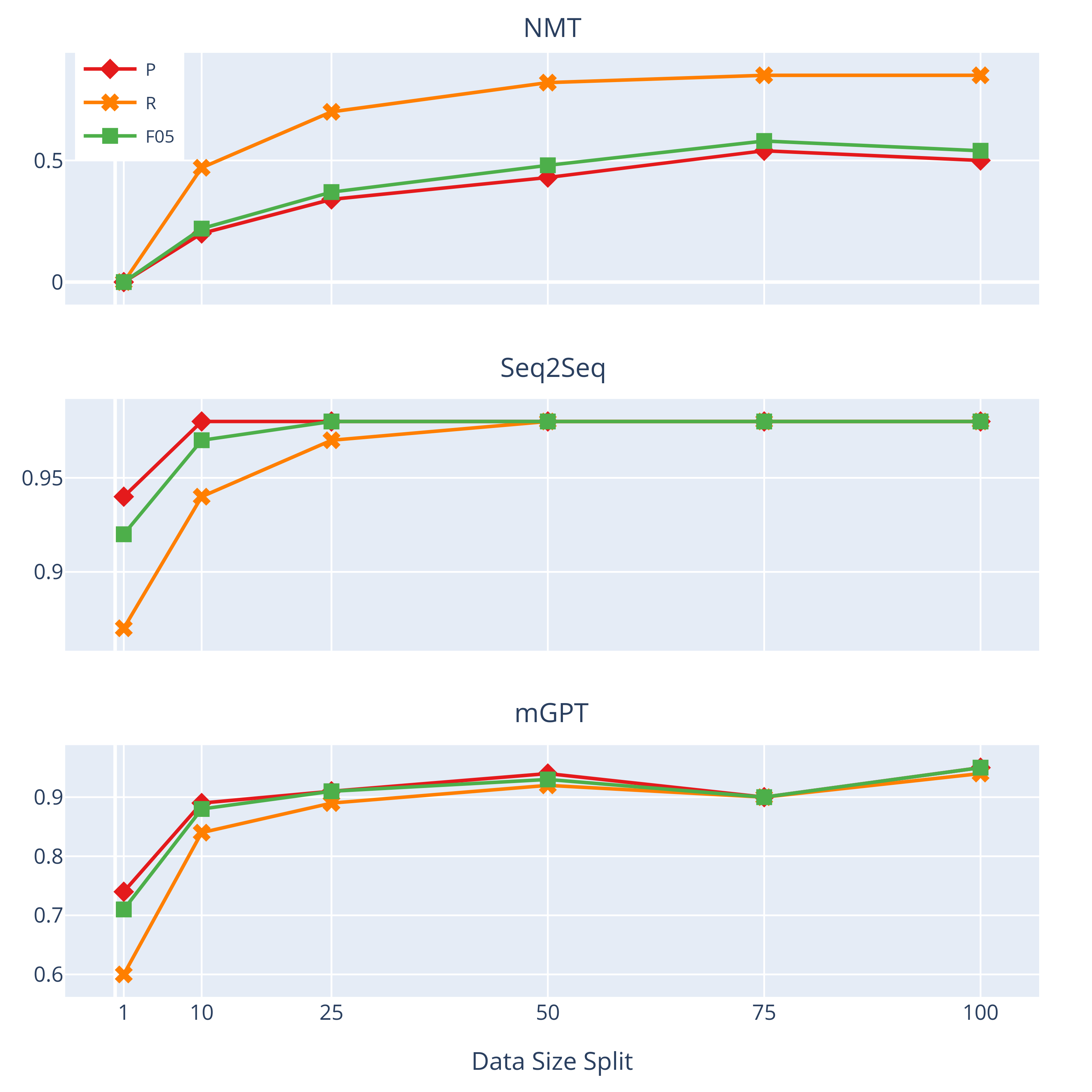}
        \caption{Correction performance of each model trained on varying sizes of GECTurk}
        \label{fig:corr-gec-dataset-sizes}
    \end{figure}

\begin{figure}[]
    \centering
    \includegraphics[width=1\linewidth]{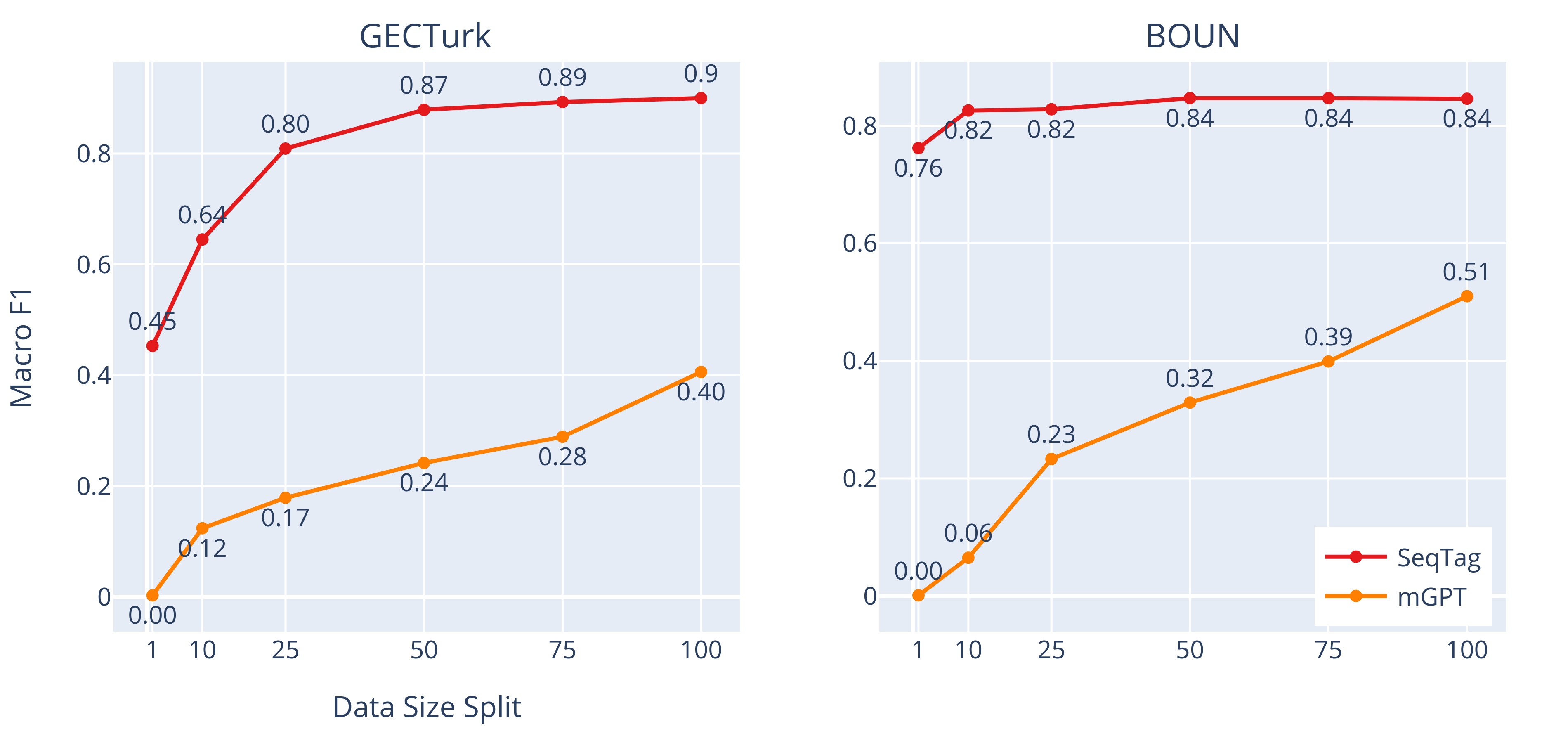}
    \caption{$F_1$ score performance of SeqTag and mGPT on GECTurk and Boun training splits.}
    \label{fig:det-full-training-dataset-sizes}
\end{figure}

    In order to obtain a better understanding of how important the dataset size is for this task, we conducted training on $1, 10, 25, 50, 75$, and $100$ percent of GECTurk and evaluated each model using the same evaluation measures. Fig.~\ref{fig:corr-gec-dataset-sizes} shows how the performance of the models vary with more training data. NMT reaches its top point with around 75\% of the training data, while SeqTag and mGPT achieve high $F_{0.5}$ scores with 25\% of the training split. However, as discussed before, correction scores can be misleadingly high, since high frequency and easy to correct errors will push the results much higher. Hence we also plot the F1 scores for the detection task both on TurkishGEC and BOUN datasets in Fig.~\ref{fig:det-full-training-dataset-sizes}. The plot shows that the GECTurk dataset is richer than the BOUN, since SeqTag and mGPT F1 performances are much steeper on the former.  

\end{document}

%% file: sections/intro.tex
Grammatical Error Correction~(GEC) is among the well-established NLP tasks with dedicated shared tasks (e.g., BEA~\cite{bryant-etal-2019-bea}), benchmarks, and even specific evaluation measures. With the increasing interest from the community, the field is in constant need of novel writing tools, and methods, and more importantly, extensions to other languages.   

Recently, there has been an explosion of research about GEC for high-resource languages, especially for English~\cite{gt5, omelianchuk-etal-2020-gector, bryant-etal-2019-bea}. These recent techniques use two main approaches: formulating the task as i) neural machine translation, i.e., generation~\cite{gt5} and ii) token classification to detect erroneous tokens~\cite{omelianchuk-etal-2020-gector}. First set of approaches mainly utilize and engineer vanilla Transformers to generate the corrected text, while, the second set focuses on engineering a set of errors and transformation rules. Nonetheless, both formulations require a large set of parallel corpora containing grammatically correct and incorrect sentence pairs. Furthermore, the latter approach additionally requires a highly curated dataset with annotations for correcting errors (i.e., location and type of the error). However, constructing such a parallel corpus with error annotations is nontrivial---especially for low-resourced languages with rich morphology like Turkish. The challenge is due to grammar rules, a.k.a., writing errors being entangled in several layers, such as phonology, morphology, syntax, and semantics. As of today, there are no spelling or grammatical error datasets, as mentioned by \citet{nlp_survey}, with the exception of the dataset introduced by \citet{detecting_clitics}. 

To address this, we focus on the Turkish Language and utilize the official writing rules established by the Turkish Language Association\footnote{\url{https://tdk.gov.tr/}}. We implement corruption, a.k.a. transformation, functions to generate instances that violate a specific rule, which requires challenging analysis of sentences on several linguistic levels, as well as curation of specialized lexicons. Then, we generate a large, synthetic, high-quality annotated corpus by applying transformation functions to professionally edited, modern Turkish articles. In addition to the transformation functions used for data generation, we implement and share the reverse-transformation functions for validating the generated datasets and developing sequence tagger models, which achieve state-of-the-art in English.

In addition, we compile a corpus of movie reviews and manually annotate 300 sentences with the proposed error types to evaluate the models in a real-life setting. Furthermore, we design and implement several baselines using standard neural machine translation~(NMT), sequence tagging and prefix-tuning. While NMT models are only trained to generate the corrected sentences, sequence tagging models are trained to tag the tokens with the error type (if any) and then perform the associated reverse transformation function on the detected error to generate the correct text. Finally, we perform prefix-tuning~\cite{Li2021PrefixTuningOC} on large multilingual language model, mGPT \citet{Shliazhko2022mGPTFL} for both detection and correction tasks to test the capacity of more recent techniques.  

Our findings indicate that our pipeline approach using smaller models perform better than employing larger pretrained models in an end-to-end fashion when it comes to both synthetic and real-world datasets---particularly for the grammatical error detection task. Conversely, we observe that pretraining benefits the models in dealing with more realistic cases, despite the larger models still falling behind their simpler counterparts. Our results from the out-of-domain tests imply that training on the synthetic dataset gives a strong prior to both smaller and larger models.

Our contributions can be summarized as follows:
\begin{itemize}
\item We propose the first comprehensive, expert-curated grammatical error schema for Turkish that covers 25 error types. 
\item We present a synthetic data generation pipeline that can be used to create arbitrary sized datasets, and can be easily extended to include new grammatical error types or lexicons, and can easily be modified to include custom tools (e.g., morphological analyzer and disambiguator).
\item We present the first large-scale, fine-grained public dataset for Turkish grammatical correction and detection, along with a manually annotated realistic test set and strong baseline models.
\end{itemize}
We make our datasets, baseline models, and synthetic data generation pipeline publicly available at \url{https://github.com/GGLAB-KU/gecturk}.

%% file: sections/relatedWork.tex
\paragraph{English GEC} Despite having a long history, with the BEA-2019 Shared Task on Grammatical Error Correction~\cite{ bryant-etal-2019-bea}, the GEC community started employing neural models and formulating GEC as a neural machine translation task (i.e., translate from grammatically incorrect to correct sentences), which has become the dominant approach. Another recent approach, GECToR~\cite{omelianchuk-etal-2020-gector}, uses the idea of reverse transformations, which can be applied to a list of source tokens $[x_{1}, …, x_{n}]$, in order to produce the desired correct grammar. Their model is a sequence tagger with a BERT encoder. Each tag corresponds to a transformation where transformations are applied after the sequence tagging finishes. In contrast, the gT5 model released by \citet{mgt5} is a multilingual mT5 model fine-tuned on artificially corrupted sentences from the mC4 corpus and uses a span prediction and classification task to fix grammatical errors~\citep{gt5}. This does require a lot of additional training time, since the original mT5 model is not initially prepared for a similar task. Their model achieves SOTA results in 4 languages while only training once.

\paragraph{Turkish GEC} Previously, \citet{detecting_clitics} proposed a neural sequence tagger model and a synthetically generated dataset to correct ``de/da'' clitic errors. In Turkish grammar, ``de/da'' is used both as a locative suffix and a conjuction meaning \textit{also}, \textit{too} that is written separately. For instance, ``-de'' is a locative suffix in the sentence ``Ev\textbf{de} (\textbf{At} home)''; while used as a conjuction here: ``Ben \textbf{de} geliyorum (I'm coming \textbf{too})''. Mistakes in using these clitics are common among native speakers, often due to some contextual subtleties and oral dialect influencing the written language. \citet{contextual_clitics} combined a contextual word embedding model, namely BERT, with a sequence tagger to correct ``de/da'' clitic errors. Although these errors are common, they constitute only a small portion of grammatical errors made by native speakers. In addition, while there are various forms of this error, the previous work only considers a few. Our data generation strategy considers multiple versions of the ``de/da'' clitic errors and many more common grammatical errors.

\paragraph{Parsing-based Approaches}
While most of the current approaches focus on reverse transformations and sequence tagging, there are several studies that involve the use of parsing techniques. \citet{more-precision-correction} create a parsing tree, and identify malformed parts of the tree to detect grammatical errors. \citet{using-rich-models} use symbolic parsers and computational grammars for GEC and GED. On the same research line, \citet{robust-parsing} use bridged analyses combined with parsing to better allow for connecting two phrases in Head-driven Phrase Structure Grammar (HPSG).

%% file: sections/methodology.tex
The overall generation process is given in Fig.~\ref{fig:pipeline}. First, we randomly sample from professionally edited Turkish corpora~( \S\ref{ssec:org_corpus}). Then, sentences are corrupted---if possible---following the expert-curated transformation rules explained in~\S\ref{ssec:trans_rules}\ggrev{}{, as well as the use of a morphological analyzer}. Finally, pairs of grammatically correct and corrupted sentences are added to the final Turkish GEC corpus following the $M^{2}$ scorer~\cite{dahlmeier-ng-2012-better} \ggrev{}{(MaxMatch)} data format.

\subsection{Corpus}
\label{ssec:org_corpus}   
    Our proposed data generation pipeline is built upon the assumption that all input sentences are grammatically correct.
    Hence, we base our study on previously compiled newspaper corpora~\cite{opinion_column_2, opinion_column_1, opinion_column_3, opinion_column_4} that are proofread and went through a professional editing process. The articles are on various topics, including politics, sports, and medicine, and have been written by more than 95 authors for three different newspapers; in total, more than 7000 singly authored documents were collected between 2004-2012.
    Once we obtained grammatically correct source sentences, we performed several preprocessing steps, such as removing duplicates (2.9\% of the combined dataset), ending with 138K unique sentences.          
\subsection{Transformation}
\label{ssec:trans_rules}
    The Turkish Language Association~(TDK)\footnote{{\url{https://www.tdk.gov.tr/}}}, a government agency founded in 1932, is responsible for providing resources to conduct scientific research on written and oral sources of Turkish. Within this scope, they specify and maintain a comprehensive list of publicly available writing rules\footnote{\url{https://www.tdk.gov.tr/kategori/icerik/yazim-kurallari/}}. We rely on this expert-curated list to generate forward and backward transformation rules, which we refer to as $f$ and $f^{-1}$ respectively. However, the list is long, and some writing rules are intuitive to native speakers, so that any errors made on these rules sound abnormal to them. \ggrev{Our criterion used for selecting writing rules most commonly mistaken among native speakers was talking to experts and narrowing down the list with their feedback. The list of selected writing rules with their associated reversible transformation functions is given in \S\ref{appendix:trans_rules}.}{We select the grammar rules that are most commonly used incorrectly by native speakers, determined by consulation with Turkish language experts and filtering the list from TDK using their feedback. We do not include any rules that are common for Turkish language learners but rarely made by native speakers.} Table \ref{table:trans_rules} provides the full list of the transformation rules produced by this work. The transformations rely on a morphological analyzer, which was essential to get the transformations right for a morphologically rich language like Turkish.  For more information on Turkish Morphology, we refer to \citet{turkish_morph_1} and \citet{turkish_morp_2}.
     
    \paragraph{Applying $f$}
    For each sampled sentence, first, we shuffle the list of $f$s. This ensures that mutually exclusive transformation functions are applied with desired frequencies. Then, we iteratively apply each $f$ on the sentence given with the pseudo-code in Algorithm~\ref{alg:pipeline}. \ggrev{A pseudo-code of the general form of these functions can be found in Algorithm~\ref{alg:pipeline}.}{} Here, $f$ gets an input sentence $\mathit{s}$, morphological analysis of the sentence $\mathit{M_s}$, an array of indicators for whether any transformation has been applied to the word--- $flags$, and parameter $\mathit{p}$ $\in$ ($0$,$1$). The algorithm, then, iterates over tokens (or pairs) and checks whether the token has been transformed. If not, it checks whether the token(s) are eligible for $f$. If eligible, we apply $f$ with the probability $\mathit{p}$, since not all errors are made with the same frequency by native speakers.\footnote{We choose the probabilities intuitively after an initial analysis on web corpus and student essays.}

    \paragraph{Eligibility Check}    
    Some official writing rules require syntactic analysis at the token and sentence levels. For instance, to apply the transformation function CONJ\_DE\_SEP, one must perform morphological analysis and disambiguation to analyze the part-of-speech tags at the morpheme level. That is, CONJ\_DE\_SEP transformation can be applied \textbf{only if} a ``-de/da'' morpheme with a \textsc{Conjuction} part-of-speech tag is found. Additionally, a small set of rules requires specialized lexicons, e.g., a list of exceptional foreign words for FOREIGN\_R2\_EXC. To address the former, we use a state-of-the-art morphological analyzer~\citet{morse_analyzer}, and the lexicons are taken from the official lists provided by TDK\footnotemark[2].

    \paragraph{Annotation Format} We use the standard GEC annotation format following \citet{ng-etal-2013-conll} and \citet{bryant-etal-2019-bea}. An example annotation is given in Fig.~\ref{fig:annot_format}. Here, S and A refer to the ungrammatical sentence and edit annotations respectively. Each A contains starting and ending indices, the error type, the corrected phrase, and the id of the annotator. 

        \alglanguage{pseudocode}
        \begin{algorithm*}
            \small
            \caption{Apply $f$}
            \label{alg:pipeline}
            \begin{algorithmic}[0]
            
            \Require $s :=$ sentence, $M_{s} :=$ morphological analysis, flags, p
            \Ensure tags
            \State $tags \gets [\,]$
            \State $n \gets$ Number of tokens in $s$
            \For {$i = 1 \to n$}
                \If {flags$[i]$ \textbf{and} $is\_eligible(s, M_s)$ \textbf{and} $flipCoin(p)$}
                    \State tags.insert("A $i$ \{$i + 1$\} $|||$ ruleID $|||$ sentence$[i]$ $|||$ REQUIRED $|||$ -NONE- $|||$ $0$")
                    \State sentence$[i] \gets$  transformed token at index i
                    \State flags$[i] \gets$ \textbf{False}
                \EndIf
            \EndFor
            \Statex
            \end{algorithmic}
              \vspace{-0.2cm}%
        \end{algorithm*}
\begin{figure}
        \centering
         \begin{subfigure}[l]{\columnwidth}
         \centering
        {\small \colorbox{gray}{Uyuyakaldığı} için hem işe gitmedi \colorbox{yellow}{hem de} akşamki yemeğe \colorbox{pink}{gelemeyecek}. \\}
        \vspace{1mm}
        
        {\scriptsize (Because they overslept, they didn't go to work and won't be able to come to dinner tonight.)}
        \caption{}
        \end{subfigure}
    
         \begin{subfigure}[l]{\columnwidth}
         \centering
        {\small \colorbox{gray}{Uyuya kaldığı} için hem işe gitmedi \colorbox{yellow}{hemde} akşamki yemeğe \colorbox{pink}{gelemiyecek}.\\}       
        \caption{}
        \end{subfigure}
    
        \vspace{1mm}
             \begin{subfigure}[l]{\columnwidth}
         \centering
        {\scriptsize S Uyuya kaldığı için hem işe gitmedi hemde akşamki yemeğe gelemiyecek. \\} 
            \vspace{0.4mm}
        {\scriptsize A 0 2|||COMP\_VERB\_ADJ|||Uyuyakaldığı|||REQUIRED|||-NONE-|||0 \\}      
                \vspace{0.4mm}
        {\scriptsize A 6 7|||CONJ\_DE\_SEP|||hem de|||REQUIRED|||-NONE-|||0 \\}
                \vspace{0.4mm}
        {\scriptsize A 9 10|||PRONOUNC\_EXC|||gelemeyecek|||REQUIRED|||-NONE-|||0 \\}  
        \caption{}
        \end{subfigure}      
        \caption{The grammatically correct sentence is given in (a), the transformed version is given in (b), and the annotation format is given in (c).}
        \label{fig:annot_format}
\end{figure}
      
\paragraph{Postprocessing}
    Despite the use of professionally edited source sentences, there are still some grammatically incorrect sentences that slip through. We detect these cases by taking advantage of a key property of our reverse transformations: since each $f$ is reversible, we should see $S = f^{-1}(f(S))$ for each sentence $S$. Therefore, at the end of the transformation process, we perform this check on every generated, grammatically incorrect sentence. If a sentence fails this check, then we know it is problematic, and we remove it from the corpus. The final sentences are thus properly modified in the desired way, with no unintentional side effects.
    
\subsection{Annotated Corpus}
\label{ssec:realcorpus}
The annotated corpus includes more than 138K sentences, with 104K error annotations belonging to 25 error types given in Table~\ref{table:trans_rules}. In this corpus, 50\% of sentences are error free, in order for models to learn how to detect/correct sentences that are already grammatically correct. The generative pipeline controls the frequency of those error types, aiming to mimic the human error frequencies (see App.~\ref{sec:errorfreq}). As in the dataset of CoNLL-2014 shared task \cite{conll_2014}, some error types appear more frequently than others. These frequencies are by the probability parameters $p$; therefore, the difference between frequencies of error types is an intended result. \ggrev{}{Our dataset is finally split into a train/val/test set of 70\%/15\%/15\%.}

\subsection{Curated Test Corpus}
\label{ssec:testcorpus}

For a more realistic test setting, we use movie reviews from a popular website~\footnote{\url{www.beyazperde.com}} shared by \citet{altinok-2023-diverse}. After performing sentence tokenization and deduplication, we use sentences that contain grammatical errors. To do so, we employ various techniques: for each rule, we utilize available dictionaries, incorporate regular expressions to identify specific morphemes that may exhibit errors, and employ models trained on a synthetic dataset. Then, our domain expert annotates the sentences for the proposed error types, following the standard GEC annotation format. By adhering to this procedure, we successfully produce a test dataset of 300 sentences, wherein half of the sentences were grammatically correct and the other half contained errors.

%% file: tables/trans_rules.tex
\def\rot{\rotatebox}

\begin{table*}
\begin{center}
\scalebox{0.7}{
\begin{tabular}{clp{0.3\linewidth}p{0.3\linewidth}p{0.2\linewidth}}

\toprule
\textbf{Category} &\textbf{Rule ID} &\textbf{Description} & \textbf{f} & \textbf{Frequency}\\
\midrule
\multirow{6}{*}[-3.5em]{\rot{90}{-DE/-DA}}& 1. CONJ\_DE\_SEP & Conjunction ``-de/-da'' is written separately. & Durumu [oğluna da -> oğlunada] bildirdi. & 12962\\
& 2. CONJ\_DE\_VH & Conjunction ``-de/-da'' must follow the vowel harmony & Çok [da -> de] iyi olmuş. & 101\\
& 3. CONJ\_DE\_AR & Conjunction ``-de/-da'' does not follow phonetic assimilation rules. & Sınıf [da -> ta] temizlendi.& 99\\
& 4. YADA & ``-de/-da'' written together with the word ``ya'' is always written separately. & Sen [ya da -> yada] o buradan gidecek. & 472\\
& 5. CONJ\_DE\_APOS & Conjunction ``-de/-da'' cannot be used with an apostrophe. & [Ayşe de -> Ayşe'de] geldi. & 10859\\
& 6. CASE\_DE & Suffix ``-de/-da'' is written adjacent. & [Evde -> Ev de] hiç süt kalmamıştı. & 37462\\
\midrule
\multirow{2}{*}[-2em]{\rot{90}{-KI}} & 7. CONJ\_KI\_SEP& Conjunction ``-ki'' is written separately. & Bugün öyle çok [yorulmuş ki -> yorulmuşki] hemen yattı. & 1817\\
& 8. CONJ\_KI\_EXC & On some exceptional instances, by convention, the conjunction ``-ki'' is written adjacent. & [Belki -> Bel ki], [oysaki -> oysa ki], [çünkü ->  çünki] & 1395\\
\midrule
\multirow{3}{*}[-2em]{\rot{90}{FOREIGN}} & 9. FOREIGN\_R1 & Words that start with double consonants of foreign origin are written without adding an ``-i'' between the letters.& [gram -> gıram] & 307\\
& 15. FOREIGN\_R2 & Some foreign origin words undergo consonant assimilation & [sebebi -> sebepi] & 327\\
& 16. FOREIGN\_R2\_EXC & Exceptions to the FOREIGN\_R2 rule & [evrakı -> evrağı] & 422\\
\midrule
\multirow{2}{*}[-1.2em]{\rot{90}{BISYL}} & 13. BISYLL\_HAPL\_VOW  & Some bisyllabic words undergo haplology when they get a suffix starting with a vowel. & [ağzı -> ağızı] & 1567\\
& 14. BISYLL\_HAPL\_VOW\_EXC & Exception to previous rule & [içeride -> içerde] & 866\\
\midrule
\multirow{2}{*}[-1em]{\rot{90}{LIGHT VERB}} & 17. LIGHT\_VERB\_SEP &  Light verbs such as ``etmek, edilmek, eylemek, olmak, olunmak'' are written separately in case of no phonological assimilation & [arz etmek -> arzetmek] & 460\\
& 18. LIGHT\_VERB\_ADJ & Light verbs are written adjacent in case of phonological assimilation e.g., liaison & [emretti -> emir etti] & 547\\
\midrule
\multirow{1}{*}[-1em]{\rot{90}{COMPOUND}} & 
20. COMP\_VERB\_ADJ & Compound words formed by knowing, giving, staying, stopping, coming, and writing are written adjacent if they have a suffix starting with -a, -e, -ı, -i, -u, -ü. & [uyuyakalma -> uyuya kalmak], [gidedurmak -> gide durmak], [çıka gelmek -> çıka gelmek] & 7840\\

\midrule
\multirow{5}{*}[-3.5em]{\rot{90}{SINGLE}} & 22. PRONOUN\_EXC & Traditionally, some pronouns are written adjacent.& [hiçbir -> hiç bir], [herhangi -> her hangi] & 3867\\
& 23. SENT\_CAP & The first letter of the sentence is capitalized.	& [Onlar -> onlar] geldi. & 2613\\
& 24. CAPPED & Some Arabic and Persian originated words are written with capped letters. & [kâğıt -> kağıt], [karargâh -> karargah] & 834\\
& 25. ABBREV & Grammatical rules for abbreviations, such as adding suffixes to abbreviations, punctuations with abbreviations etc. & [Alm. -> Alm], [THY'de -> THY'da], [cm'yi -> cm'ye] & 359\\
& 12. PRONOUNC\_EXC & Unlike its pronunciation, verbs ending with ``-a/-e'' do not mutate when they get a suffix other than ``-yor'' & [başlayacağım -> başlıyacağım] & 12750\\
\bottomrule
\end{tabular}
}
\caption{Selected list of writing rules introduced by Turkish Language Association. $f[arg1 -> arg2]$ refers to the transformation function where the correct and corrupted surface forms are given with $arg1$ and $arg2$ respectively.}
\label{table:trans_rules}
\end{center}
\end{table*}

%% file: sections/models.tex
In this paper, we consider two tasks: Grammatical Error Correction~(GEC) and Grammatical Error Detection~(GED). 
\paragraph{Grammatical Error Correction~(GEC)} takes as input a grammatically incorrect sentence and outputs the corrected version of the sentence. Formally, given an input sentence $\textbf{x} = (x_{1}, \cdots, x_{T})$ which may contain some grammar mistakes, the aim is to produce an output sentence $\textbf{y} = (y_{1}, \cdots, y_{T'})$ which contains no grammatical errors. Conditions are not imposed on how the model produces grammatically correct sentences.

\paragraph{Grammatical Error Detection~(GED)} takes a slightly different approach to this problem, with the goal of producing detailed information about the errors in the source sentence. This includes details about the type of error and the location of the error in the sentence. Formally, given an input sentence $\textbf{x} = (x_{1}, \cdots, x_{T})$, we can represent the problem as a token-level classification task, where the output is $\textbf{c} = (c_{1}, \cdots, c_{T})$, and $c_{i}$ represents the error type of token $i$. Given the knowledge that an error of type $c_{k}$ occurred at the location from $m$ to $n$, it is then possible to apply the corresponding reverse transformation $f^{-1}$, and fix the error.

\subsection{Models}
We introduce three models to evaluate the performance of GECTurk: An NMT baseline, a sequence tagger using BERT~\cite{Devlin2019BERTPO} pretrained on Turkish, and mGPT using prefix-tuning. All models are trained using 1 Nvidia V100 GPU. We only provide the essential information about the models here. More details are available in Appendix \ref{app:models}.

\paragraph{NMT Baseline:} We train a vanilla transformer model \cite{Vaswani2017AttentionIA} for GEC. This choice is inspired by the most recent shared task on grammatical correction~\cite{bryant-etal-2019-bea}, where many of the winning teams used transformer-based models and modeled the problem as a Neural Machine Translation (NMT). The training dataset consists of triples $\{(x_{i}, y_{i}, a_{i})\}_{i=1}^{N}$, where $x_{i}$ is the $i$-th input sentence, $y_{i}$ is the corresponding ground truth corrected sentence, and $a_{i}$ are the annotations. During training, the model receives $x_{i}$ as input. Due to the nature of the formulation, NMT is only used for correction. 

\paragraph{Sequence Tagger:} Similar to recent work~\cite{omelianchuk-etal-2020-gector}, we train a sequence tagging model using a cased BERT encoder, pretrained on Turkish text \cite{BERTurk} with default configurations and additional linear and softmax layers on the top. The BERT model uses the WordPiece tokenizer \cite{wordpiece_tokenizer} that segments tokens into subwords. Therefore, each sentence in the dataset is first tokenized into subwords and passed into the BERT encoder. We only hold the first subword's representation for words with multiple subword tokens. Then, the encoder's representations are linearly transformed and passed to the softmax layer to classify into possible error types described in Table~\ref{table:trans_rules} or no error. The model is finetuned for token classification objective using cross-entropy loss. The loss is similar to Eq.~\ref{eq1} located in the Appendix, except ranging over the number of possible error types. 
 
The advantage of this model is the ability to perform error detection easily, as opposed to simply error correction. Correction is simply performed with reverse transformations as described previously.

\paragraph{Prefix Tuning:} Inspired by the recent successes of prefix tuning~\cite{Li2021PrefixTuningOC} as an alternative to model fine-tuning, we use OpenPrompt~\cite{Ding2021OpenPromptAO} to perform prefix tuning on mGPT~\cite{Shliazhko2022mGPTFL}. Despite being multilingual and primarily focused on other languages, mGPT achieves encouraging results on morphologically rich languages~\cite{sahinMorph22}. In prefix tuning, we append $N$ trainable (soft) tokens to the front of each input. Therefore, given input $\textbf{x} = (x_{1}, \cdots, x_{T})$, the new input becomes $\textbf{x} = (s_{1}, \cdots, s_{N}, x_{1}, \cdots, x_{T})$, where the $s_{i}$'s are the added artificial tokens. We then optimize only these tokens during training, while leaving the original model frozen. We primarily utilize mGPT for sequence generation, where the model simply outputs the grammatically correct sentence. We use the standard sequence generation prompt provided by OpenPrompt, due to its recent success~\cite{sahinMorph22}, and use teacher forcing during training. When the model detects multiple types of errors in the same sentence, all the detection information is generated on the same line. 

\begin{figure}
    \centering
    \includegraphics[scale=0.75]{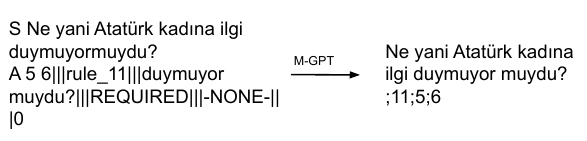}
    \caption{Example output of the mGPT model. The first line performs the grammatical error correction, and subsequent lines allow for detection. The annotations are included for comparison with model outputs, but are not actually provided to the model.}
    \label{fig:mgpt-example}
\end{figure}

Here, we model both correction and detection tasks in the same sequence generation approach, where the corrected sentence is first generated, and then information about the violated rule, and the location of this error is generated at the end of the sentence. This allows for one trained model to output both results. In order to train this correctly, the target sentence was appended with the details of the error type and location, and used for loss calculations. An example is provided in Fig~\ref{fig:mgpt-example}.

%% file: sections/setup.tex
\subsection{Datasets}
\label{ssec:dataset}

\begin{table}[!t]
\centering
\scalebox{0.7}{
\begin{tabular}{lrrrr} 
\toprule
\textbf{Dataset}   & \textbf{\#s} & \textbf{\#a} & \begin{tabular}[c]{@{}c@{}}\textbf{\%e}\end{tabular} & \begin{tabular}[c]{@{}c@{}}\textbf{\#e}\end{tabular}  \\ 
\midrule
(Ours) GECTurk & 138K  & 104K   & 49.7\%  & 25 \\
(Ours) CuratedTest & 300  & 227  & 50\%  & 25 \\
BOUN  & 10K & 6K  & 50.0\% & 2 \\
BOUN complex  & 102  & 105  & 100.0\%  & 2  \\
\cline{1-2}\cline{3-3}\cline{4-5}
\end{tabular}
}
\caption{Datasets used for training and evaluation. \#s: sentences, \#a: annotations, \%e: percentage of erroneous sentences, \#e: number of errors types}
\label{table:dataset_info}
\end{table}
The list of datasets and their statistics are given in Table~\ref{table:dataset_info}. GECTurk and MovieReview datasets are already described in \S\ref{ssec:realcorpus} and \S\ref{ssec:testcorpus} accordingly. The BOUN dataset~\cite{detecting_clitics} is a relatively smaller dataset of 15K training and 2K test sentences, containing only $2$ error types. It also includes a complex split, a list of $100$ sentences that are mentioned to be extra challenging by the authors. 

\subsection{Evaluation}
\label{ssec:eval}

\paragraph{Grammatical Error Correction}
Following the \citet{omelianchuk-etal-2020-gector} and \citet{bryant-etal-2019-bea}, we report Precision ($P$), Recall ($R$), and $F_{0.5}$ scores using the $M^{2}$~(MaxMatch) scorer~\cite{dahlmeier-ng-2012-better}.

\paragraph{Grammatical Error Detection}
To allow for a fair comparison with the BOUN dataset from \citet{detecting_clitics}, we use the same metrics, namely Precision ($P$), Recall ($R$), and $F_{1}$. Since the task is modeled as a sequence tagging problem, this aligns with the standard evaluation for sequence tagging, such as in \citet{Huang2015BidirectionalLM}. For all GED results, we report macro metrics, which are computed by taking unweighted average of each classes' result. We use macro metrics over micro ones since distribution of grammatical errors types made by humans are imbalanced. To calculate these scores, we use SeqEval~\cite{seqeval}, a common library for evaluating sequence tagging tasks and use one tag for each error type.

%% file: sections/results.tex
We conduct a set of experiments to investigate the performance of the proposed baselines and the difficulty level of the introduced dataset: GECTurk. To do so, we train the baseline models using the experimental setup explained in Section~\S\ref{sec:setup} using three different fixed seeds. The mean and standard deviation of their performances on both GEC and GED (if applicable) are given in Table~\ref{table:main_results}. As can be seen, both SeqTag and mGPT provide exceptionally strong results over $0.94$ $F_{0.5}$ score for the GEC task, compared to the NMT baseline model. On the other hand, the detection task is performed more competently by SeqTag---as expected---than mGPT, which again achieves around $0.90$ $F_{1}$ score. Moreover, the experiment on the effect of dataset sizes shows that the proposed dataset is indeed more challenging than existing ones, as the models demonstrate a steeper learning curve due to the larger number of error types. More information on the dataset size experiments can be found in Appendix~\ref{appendix:dataset_size}. 

The reason why successful detection does not always translate into successful correction is because erroneous edits undermine the grammatical accuracy of the sentence, even when the grammatical error is successfully detected. Here, the $M^2$ scorer effectively identifies this anomaly and explains the decrease in the correction scores. It is also worth noting that GED and GEC are two separate tasks, and models handle them differently. For example, generative models, such as mGPT, generate both the corrections and detections by predicting the next tokens, so there isn't necessarily a strong correlation between what is generated for each of them. On the other hand, SeqTag uses a pipeline approach, where it detects the errors first, then applies reverse transformations to fix them. Hence there is a stronger correlation between the detection and correction performances for SeqTag, as expected. 

In order to test whether the performance of our models would transfer to different domains, we perform zero-shot experiments on the curated test set from the movie domain. We use the checkpoints that performed highest on our synthetic test set, and evaluate without any additional training on the hand-annotated corpus as given in Table \ref{table:main_results}, second row. For the detection task, SeqTag performs similarly to synthetic setting, while mGPT's performance \textit{increases dramatically}, proving the importance of being exposed to real-life data during pretraining. However, SeqTag still outperforms mGPT by a large margin due to its classification objective. On the other hand, both models perform significantly worse on the correction task for the currated test data compared to the synthetic setting, suggesting a larger room for improvement on this more challenging test set. 

\subsection{Knowledge Transfer}

\begin{table}[]
    \centering
    \scalebox{0.48}{
    \resizebox{\textwidth}{!}{
    \begin{threeparttable}
    \begin{tabular}{cccccccc}
        \toprule
        \multicolumn{8}{c}{\textbf{BOUN Zero-Shot}}\\
        & \multicolumn{4}{c}{\textbf{Detection}} & \multicolumn{3}{c}{\textbf{Correction}}\\
        \cmidrule(lr){2-5} \cmidrule(lr){6-8}
        &P &R & $F_{1}$ &Acc &P &R & $\mathbf{F_{0.5}}$\\
        \midrule
        NMT   &-  &- &- &- &0.19 &0.56 &0.22 \\
        SeqTag &0.91 &0.72 &0.80 &0.99 &0.81 &0.63 &0.77 \\
        mGPT   &0.60 &0.56 &0.58 &0.97 &0.75 &0.67 &0.73 \\       
        \midrule
        \multicolumn{8}{c}{\textbf{BOUN Complex Split Zero-Shot}}\\
        \midrule
        NMT    &- &- &- &- &0.71 &0.73 &0.72 \\
        SeqTag &0.99 &0.93 &0.96 &0.99 &0.99 &0.93 & 0.98\\
        mGPT   &0.61 &0.96 &0.90 &0.93 &0.79 &0.75 &0.78 \\       
        \midrule
        \multicolumn{8}{c}{\textbf{BOUN Full Training}}\\
        \midrule
        NMT    &- &- &- &- &0 &0 &0 \\
        SeqTag &0.97 &0.86 &0.91 &0.99 &0.97 &0.86 & 0.94\\
        mGPT   &0.61 &0.48 &0.53 &0.97 &0.85 &0.73 &0.82 \\
        \toprule
        BOUN Results &0.92 &0.82 &0.87 &0.71 \tnote{*}
        &- &- &- \\
        \bottomrule
    \end{tabular}
    \begin{tablenotes}
        \item[*] This result is for BOUN Complex Split Zero-Shot task.
    \end{tablenotes}
    \end{threeparttable}
    }
    }
    \caption{Performance metrics of various models on the BOUN dataset. The table is divided into three sections: models trained on the GECTurk dataset and evaluated zero-shot on two different BOUN splits, and models exclusively trained and evaluated on BOUN.}
    \label{table:all_results}
\end{table}

Next, we investigate the transfer capacity of our models on unseen datasets using a different set of errors~(i.e., mostly a subset) originally introduced to our models. We first evaluate our pretrained models on the BOUN~\cite{detecting_clitics} standard and complex test splits to gain insights into their zero-shot ability, given with Table~\ref{table:all_results}, first row. Surprisingly, our best model, SeqTag, achieves 0,80 $F_{1}$ that is on-par with state-of-the-art for the standard split. It also surpasses state-of-the-art accuracy scores on the complex split by a large margin together with the mGPT model. This result suggests that, the error type knowledge is mostly transferrable to other domains. Similar to our results on GECTurk, mGPT scores considerably low on detection compared to SeqTag. However, the performance of mGPT is higher on BOUN dataset due to the small number of error types that are relatively more balanced. We note that despite the claims made by the authors of the BOUN dataset, our results suggest no additional complexity in the ``complex'' split as shown in Table~\ref{table:all_results}, second row. 

Finally, we investigate the effectiveness of our general approach by training our proposed models from scratch on the BOUN~\cite{detecting_clitics} training split, given in Table~\ref{table:all_results}, BOUN Full Training. For this setup, the NMT model was not able to fully converge, and just produced unhelpful noise, hence, shown as 0.

Following our previous results, SeqTag achieves $F_{1}$ score of $0.91$, surpassing the state-of-the-art by $0.04$ pp, and its zero-shot performance by a large margin ($0.11$ pp). This suggests that, pipeline approach is able to transfer a considerable amount of knowledge, however, there is still a large gap that can be compensated by directly training on the actual domain and error types. On the other hand, the high scores provide cues for the strength of the proposed model. Surprisingly, prefix tuning of mGPT model directly on the BOUN training dataset does not increase the performance compared to the zero-shot setting. This suggests two things: i) synthetic dataset such as ours, GECTurk, provides quality prior knowledge on the Turkish grammatical error types and ii) pretrained models have a considerably larger transfer capacity compared to training from scratch, as expected. Furthermore, for languages where the error types are mostly identified and can be fixed by a set of rules, a pipeline approach such as SeqTag proves more effective, efficient and robust. 

%% file: sections/conclusion.tex
In this work, we have presented an annotated dataset for Grammatical Error Correction (GEC) and Detection (GED), GECTurk, containing more than 20 Turkish writing error types proposed by Turkish language experts. We have also introduced a flexible and extensible data generation pipeline that can be used to create a synthetic dataset from grammatically correct sentences. We used this pipeline to create a large-scale dataset using multiple opinion columns from Turkish newspapers. In addition, we have manually constructed a more challenging test set by annotating the movie-reviews with the proposed error types. 

Finally, we implemented a diverse set of strong baseline models, by training from scratch, fine-tuning, or if applicable, using prefix tuning. Our results show that simpler models focusing on the smaller problem of detecting the error types outperform large pretrained models on both the synthetic and real-life datasets, especially for the detection task. On the other hand, we observe that pretraining helps the models to handle more realistic cases, even though they still lag behind the simpler models. Our out-of-domain results suggest that training on the synthetic data gives a strong prior to both smaller and larger models.  

%% file: sections/limitations.tex
There are a few key limitations to our work. One key issue is that mGPT is very computationally intensive to work with, even when only doing prefix tuning. This prevented us from training until fully converged, and instead only opted for 1 epoch. Another limitation is that the data generation pipeline is very time-consuming as it requires the use of a morphological analyzer. This prevents the pipeline from being used for very large-scale datasets. Additionally, the requirements for hand-crafted rules and reverse transformations slows the speed at which new rules can be added, and which rules can even be added. 

\ggrev{}{Another important limitation is the necessity of dictionaries to handle exceptions to grammatical rules. Words in Turkish that have been borrowed from other languages (notably Persian, French, and Arabic) tend to not align with the normal grammar rules, and thus require special lists of exceptions. While we aimed to include as many as possible, it is definitely possible that we missed some, which can lead to rare edge cases where our pipeline fails. While dictionaries do allow for incorporating learned knowledge directly into the process, and is certainly an invaluable part of our pipeline, these edge cases can cause problems during dataset generation, so should be considered a limitation.}

%% file: sections/ethics.tex
Transformer-based large language models such as \cite{Vaswani2017AttentionIA, Shliazhko2022mGPTFL, gt5} are very successful, but they also have some ethical concerns. First, the model is highly dependent on the dataset it was pre-trained on. It is possible that the dataset contained certain biases, such as racism or sexism, which will later be passed to the model outputs. While this is an important detail to focus on, our work does not focus much on actually training these models. We perform training using GECTurk, which is collected from opinion columns, as well as the BOUN dataset, which is already publicly available. Due to the publication of these columns in reputable Turkish newspapers, they contain less bias than the average document collected online. While the pre-training procedure itself may have introduced some biases, we are unable to handle these in our work. \ggrev{}{While these problems are inherent to all deep learning models, we emphasize transformer models here due to our model choices.}

Another possible ethical issue is the misuse of grammatical correction models for cheating. By having a model that can automatically detect and correct grammatical errors, students can more easily use these to score better than normal on assignments and exams, with less time invested. Not only is this bad for the student's learning, but it also affects others who can be negatively impacted by the student's artificial success. 

Despite the concerns about misuse or inherent biases in the model, we believe that grammatical error correction models are more beneficial than harmful. Many people, from authors and writers, to language learners, can benefit from having grammar corrections. By introducing a dataset and demonstrating models on Turkish, an under-served language in the NLP community, more people will be able to take advantage of this, similar to the many existing tools for English. 

%% file: acl_latex.bbl
\begin{thebibliography}{33}
\expandafter\ifx\csname natexlab\endcsname\relax\def\natexlab#1{#1}\fi

\bibitem[{Acikgoz et~al.(2022)Acikgoz, Chubakov, Kural, {\c{S}}ahin, and
  Yuret}]{sahinMorph22}
Emre~Can Acikgoz, Tilek Chubakov, Muge Kural, G{\"o}zde {\c{S}}ahin, and Deniz
  Yuret. 2022.
\newblock \href {https://doi.org/10.18653/v1/2022.mrl-1.10} {Transformers on
  multilingual clause-level morphology}.
\newblock In \emph{Proceedings of the The 2nd Workshop on Multi-lingual
  Representation Learning (MRL)}, pages 100--105, Abu Dhabi, United Arab
  Emirates (Hybrid). Association for Computational Linguistics.

\bibitem[{Altinok(2023)}]{altinok-2023-diverse}
Duygu Altinok. 2023.
\newblock \href {https://doi.org/10.18653/v1/2023.acl-long.768} {A diverse set
  of freely available linguistic resources for {T}urkish}.
\newblock In \emph{Proceedings of the 61st Annual Meeting of the Association
  for Computational Linguistics (Volume 1: Long Papers)}, pages 13739--13750,
  Toronto, Canada. Association for Computational Linguistics.

\bibitem[{Amasyal{\i} and Diri(2006)}]{opinion_column_1}
M.~Fatih Amasyal{\i} and Banu Diri. 2006.
\newblock Automatic {T}urkish text categorization in terms of author, genre and
  gender.
\newblock In \emph{Natural Language Processing and Information Systems}, pages
  221--226, Berlin, Heidelberg. Springer Berlin Heidelberg.

\bibitem[{Arikan et~al.(2019)Arikan, G{\"{u}}ng{\"{o}}r, and
  Uskudarli}]{detecting_clitics}
Ugurcan Arikan, Onur G{\"{u}}ng{\"{o}}r, and Suzan Uskudarli. 2019.
\newblock \href {https://doi.org/10.26615/978-954-452-056-4\_009} {Detecting
  clitics related orthographic errors in {T}urkish}.
\newblock In \emph{Proceedings of the International Conference on Recent
  Advances in Natural Language Processing, {RANLP} 2019, Varna, Bulgaria,
  September 2-4, 2019}, pages 71--76. {INCOMA} Ltd.

\bibitem[{Bryant et~al.(2019)Bryant, Felice, Andersen, and
  Briscoe}]{bryant-etal-2019-bea}
Christopher Bryant, Mariano Felice, {\O}istein~E. Andersen, and Ted Briscoe.
  2019.
\newblock \href {https://doi.org/10.18653/v1/W19-4406} {The {BEA}-2019 shared
  task on grammatical error correction}.
\newblock In \emph{Proceedings of the Fourteenth Workshop on Innovative Use of
  NLP for Building Educational Applications}, pages 52--75, Florence, Italy.
  Association for Computational Linguistics.

\bibitem[{Can and Amasyalı(2016)}]{opinion_column_3}
Ender Can and Mehmet~Fatih Amasyalı. 2016.
\newblock \href {https://doi.org/10.1109/SIU.2016.7495711} {Text2arff: A text
  representation library}.
\newblock In \emph{2016 24th Signal Processing and Communication Application
  Conference (SIU)}, pages 197--200.

\bibitem[{{\c{C}}{\"{o}}ltekin et~al.(2023){\c{C}}{\"{o}}ltekin,
  Do{\u{g}}ru{\"{o}}z, and {\c{C}}etinoglu}]{nlp_survey}
{\c{C}}agri {\c{C}}{\"{o}}ltekin, A.~Seza Do{\u{g}}ru{\"{o}}z, and {\"{O}}zlem
  {\c{C}}etinoglu. 2023.
\newblock \href {https://doi.org/10.1007/s10579-022-09625-0} {Correction to:
  Resources for {T}urkish natural language processing: {A} critical survey}.
\newblock \emph{Lang. Resour. Evaluation}, 57(1):489.

\bibitem[{da~Costa(2021)}]{using-rich-models}
Lu{\'i}s~Morgado da~Costa. 2021.
\newblock \href {https://api.semanticscholar.org/CorpusID:247266314}
  {\emph{Using rich models of language in grammatical error detection}}.
\newblock Ph.D. thesis, Nanyang Technological University.

\bibitem[{Dahlmeier and Ng(2012)}]{dahlmeier-ng-2012-better}
Daniel Dahlmeier and Hwee~Tou Ng. 2012.
\newblock \href {https://aclanthology.org/N12-1067} {Better evaluation for
  grammatical error correction}.
\newblock In \emph{Proceedings of the 2012 Conference of the North {A}merican
  Chapter of the Association for Computational Linguistics: Human Language
  Technologies}, pages 568--572, Montr{\'e}al, Canada. Association for
  Computational Linguistics.

\bibitem[{Dayanik et~al.(2018)Dayanik, Aky{\"{u}}rek, and
  Yuret}]{morse_analyzer}
Erenay Dayanik, Ekin Aky{\"{u}}rek, and Deniz Yuret. 2018.
\newblock \href {http://arxiv.org/abs/1805.07946} {Morphnet: {A}
  sequence-to-sequence model that combines morphological analysis and
  disambiguation}.
\newblock \emph{CoRR}, abs/1805.07946.

\bibitem[{Devlin et~al.(2019)Devlin, Chang, Lee, and
  Toutanova}]{Devlin2019BERTPO}
Jacob Devlin, Ming{-}Wei Chang, Kenton Lee, and Kristina Toutanova. 2019.
\newblock \href {https://doi.org/10.18653/v1/n19-1423} {{BERT:} pre-training of
  deep bidirectional transformers for language understanding}.
\newblock In \emph{Proceedings of the 2019 Conference of the North American
  Chapter of the Association for Computational Linguistics: Human Language
  Technologies, {NAACL-HLT} 2019, Minneapolis, MN, USA, June 2-7, 2019, Volume
  1 (Long and Short Papers)}, pages 4171--4186. Association for Computational
  Linguistics.

\bibitem[{Ding et~al.(2022)Ding, Hu, Zhao, Chen, Liu, Zheng, and
  Sun}]{Ding2021OpenPromptAO}
Ning Ding, Shengding Hu, Weilin Zhao, Yulin Chen, Zhiyuan Liu, Haitao Zheng,
  and Maosong Sun. 2022.
\newblock \href {https://doi.org/10.18653/v1/2022.acl-demo.10} {Openprompt: An
  open-source framework for prompt-learning}.
\newblock In \emph{Proceedings of the 60th Annual Meeting of the Association
  for Computational Linguistics, {ACL} 2022 - System Demonstrations, Dublin,
  Ireland, May 22-27, 2022}, pages 105--113. Association for Computational
  Linguistics.

\bibitem[{Diri and Amasyali(2003)}]{opinion_column_2}
Banu Diri and Mehmet~Fatih Amasyali. 2003.
\newblock Automatic author detection for {T}urkish texts.

\bibitem[{Flickinger and Packard()}]{robust-parsing}
Dan Flickinger and Woodley Packard.
\newblock Robust parsing in hpsg: Bridging the coverage chasm.
\newblock Poster presented at the 22nd International Conference on HPSG. 2015.

\bibitem[{Flickinger and Yu(2013)}]{more-precision-correction}
Dan Flickinger and Jiye Yu. 2013.
\newblock \href {https://api.semanticscholar.org/CorpusID:10479821} {Toward
  more precision in correction of grammatical errors}.
\newblock In \emph{CoNLL Shared Task}.

\bibitem[{Huang et~al.(2015)Huang, Xu, and Yu}]{Huang2015BidirectionalLM}
Zhiheng Huang, Wei Xu, and Kai Yu. 2015.
\newblock \href {http://arxiv.org/abs/1508.01991} {Bidirectional {LSTM-CRF}
  models for sequence tagging}.
\newblock \emph{CoRR}, abs/1508.01991.

\bibitem[{{Kemik NLP Group}(2022)}]{opinion_column_4}
{Kemik NLP Group}. 2022.
\newblock Our datasets.
\newblock \url{http://www.kemik.yildiz.edu.tr/data/File/2500koseyazisi.rar}.
\newblock Online; accessed 26-November-2022]".

\bibitem[{Kingma and Ba(2014)}]{Kingma2014AdamAM}
Diederik~P. Kingma and Jimmy Ba. 2014.
\newblock Adam: A method for stochastic optimization.
\newblock \emph{CoRR}, abs/1412.6980.

\bibitem[{Lewis(1985)}]{turkish_morp_2}
G.~Lewis. 1985.
\newblock \emph{Turkish Grammar}.
\newblock Oxford University Press.

\bibitem[{Li and Liang(2021)}]{Li2021PrefixTuningOC}
Xiang~Lisa Li and Percy Liang. 2021.
\newblock Prefix-tuning: Optimizing continuous prompts for generation.
\newblock \emph{Proceedings of the 59th Annual Meeting of the Association for
  Computational Linguistics and the 11th International Joint Conference on
  Natural Language Processing (Volume 1: Long Papers)}, abs/2101.00190.

\bibitem[{Loshchilov and Hutter(2019)}]{adamw}
Ilya Loshchilov and Frank Hutter. 2019.
\newblock \href {https://openreview.net/forum?id=Bkg6RiCqY7} {Decoupled weight
  decay regularization}.
\newblock In \emph{7th International Conference on Learning Representations,
  {ICLR} 2019, New Orleans, LA, USA, May 6-9, 2019}. OpenReview.net.

\bibitem[{Nakayama(2018)}]{seqeval}
Hiroki Nakayama. 2018.
\newblock \href {https://github.com/chakki-works/seqeval} {{seqeval}: A python
  framework for sequence labeling evaluation}.
\newblock Software available from https://github.com/chakki-works/seqeval.

\bibitem[{Ng et~al.(2014)Ng, Wu, Briscoe, Hadiwinoto, Susanto, and
  Bryant}]{conll_2014}
Hwee~Tou Ng, Siew~Mei Wu, Ted Briscoe, Christian Hadiwinoto, Raymond~Hendy
  Susanto, and Christopher Bryant. 2014.
\newblock \href {https://doi.org/10.3115/v1/W14-1701} {The {C}o{NLL}-2014
  shared task on grammatical error correction}.
\newblock In \emph{Proceedings of the Eighteenth Conference on Computational
  Natural Language Learning: Shared Task}, pages 1--14, Baltimore, Maryland.
  Association for Computational Linguistics.

\bibitem[{Ng et~al.(2013)Ng, Wu, Wu, Hadiwinoto, and
  Tetreault}]{ng-etal-2013-conll}
Hwee~Tou Ng, Siew~Mei Wu, Yuanbin Wu, Christian Hadiwinoto, and Joel Tetreault.
  2013.
\newblock \href {https://aclanthology.org/W13-3601} {The {C}o{NLL}-2013 shared
  task on grammatical error correction}.
\newblock In \emph{Proceedings of the Seventeenth Conference on Computational
  Natural Language Learning: Shared Task}, pages 1--12, Sofia, Bulgaria.
  Association for Computational Linguistics.

\bibitem[{Oflazer(2014)}]{turkish_morph_1}
Kemal Oflazer. 2014.
\newblock \href {https://doi.org/10.1007/s10579-014-9267-2} {Turkish and its
  challenges for language processing}.
\newblock \emph{Lang. Resour. Evaluation}, 48(4):639--653.

\bibitem[{Omelianchuk et~al.(2020)Omelianchuk, Atrasevych, Chernodub, and
  Skurzhanskyi}]{omelianchuk-etal-2020-gector}
Kostiantyn Omelianchuk, Vitaliy Atrasevych, Artem Chernodub, and Oleksandr
  Skurzhanskyi. 2020.
\newblock \href {https://doi.org/10.18653/v1/2020.bea-1.16} {{GECT}o{R} {--}
  grammatical error correction: Tag, not rewrite}.
\newblock In \emph{Proceedings of the Fifteenth Workshop on Innovative Use of
  NLP for Building Educational Applications}, pages 163--170, Seattle, WA, USA
  → Online. Association for Computational Linguistics.

\bibitem[{Rothe et~al.(2021)Rothe, Mallinson, Malmi, Krause, and Severyn}]{gt5}
Sascha Rothe, Jonathan Mallinson, Eric Malmi, Sebastian Krause, and Aliaksei
  Severyn. 2021.
\newblock \href {https://doi.org/10.18653/v1/2021.acl-short.89} {A simple
  recipe for multilingual grammatical error correction}.
\newblock In \emph{Proceedings of the 59th Annual Meeting of the Association
  for Computational Linguistics and the 11th International Joint Conference on
  Natural Language Processing (Volume 2: Short Papers)}, pages 702--707,
  Online. Association for Computational Linguistics.

\bibitem[{Schweter(2020)}]{BERTurk}
Stefan Schweter. 2020.
\newblock \href {https://doi.org/10.5281/zenodo.3770924} {Berturk - bert models
  for {T}urkish}.

\bibitem[{Shliazhko et~al.(2022)Shliazhko, Fenogenova, Tikhonova, Mikhailov,
  Kozlova, and Shavrina}]{Shliazhko2022mGPTFL}
Oleh Shliazhko, Alena Fenogenova, Maria Tikhonova, Vladislav Mikhailov,
  Anastasia Kozlova, and Tatiana Shavrina. 2022.
\newblock \href {https://doi.org/10.48550/arXiv.2204.07580} {m{GPT}: Few-shot
  learners go multilingual}.
\newblock \emph{CoRR}, abs/2204.07580.

\bibitem[{Vaswani et~al.(2017)Vaswani, Shazeer, Parmar, Uszkoreit, Jones,
  Gomez, Kaiser, and Polosukhin}]{Vaswani2017AttentionIA}
Ashish Vaswani, Noam Shazeer, Niki Parmar, Jakob Uszkoreit, Llion Jones,
  Aidan~N Gomez, \L~ukasz Kaiser, and Illia Polosukhin. 2017.
\newblock \href
  {https://proceedings.neurips.cc/paper/2017/file/3f5ee243547dee91fbd053c1c4a845aa-Paper.pdf}
  {Attention is all you need}.
\newblock In \emph{Advances in Neural Information Processing Systems},
  volume~30. Curran Associates, Inc.

\bibitem[{Wu et~al.(2016)Wu, Schuster, Chen, Le, Norouzi, Macherey, Krikun,
  Cao, Gao, Macherey, Klingner, Shah, Johnson, Liu, Kaiser, Gouws, Kato, Kudo,
  Kazawa, Stevens, Kurian, Patil, Wang, Young, Smith, Riesa, Rudnick, Vinyals,
  Corrado, Hughes, and Dean}]{wordpiece_tokenizer}
Yonghui Wu, Mike Schuster, Zhifeng Chen, Quoc~V. Le, Mohammad Norouzi, Wolfgang
  Macherey, Maxim Krikun, Yuan Cao, Qin Gao, Klaus Macherey, Jeff Klingner,
  Apurva Shah, Melvin Johnson, Xiaobing Liu, Lukasz Kaiser, Stephan Gouws,
  Yoshikiyo Kato, Taku Kudo, Hideto Kazawa, Keith Stevens, George Kurian,
  Nishant Patil, Wei Wang, Cliff Young, Jason Smith, Jason Riesa, Alex Rudnick,
  Oriol Vinyals, Greg Corrado, Macduff Hughes, and Jeffrey Dean. 2016.
\newblock \href {http://arxiv.org/abs/1609.08144} {Google's neural machine
  translation system: Bridging the gap between human and machine translation}.
\newblock \emph{CoRR}, abs/1609.08144.

\bibitem[{Xue et~al.(2021)Xue, Constant, Roberts, Kale, Al-Rfou, Siddhant,
  Barua, and Raffel}]{mgt5}
Linting Xue, Noah Constant, Adam Roberts, Mihir Kale, Rami Al-Rfou, Aditya
  Siddhant, Aditya Barua, and Colin Raffel. 2021.
\newblock \href {https://doi.org/10.18653/v1/2021.naacl-main.41} {m{T}5: A
  massively multilingual pre-trained text-to-text transformer}.
\newblock In \emph{Proceedings of the 2021 Conference of the North American
  Chapter of the Association for Computational Linguistics: Human Language
  Technologies}, pages 483--498, Online. Association for Computational
  Linguistics.

\bibitem[{Öztürk et~al.(2020)Öztürk, Değirmenci, Güngör, and
  Uskudarli}]{contextual_clitics}
Hasan Öztürk, Alperen Değirmenci, Onur Güngör, and Suzan Uskudarli. 2020.
\newblock \href {https://doi.org/10.1109/SIU49456.2020.9302477} {The role of
  contextual word embeddings in correcting the ‘de/da’ clitic errors in
  {T}urkish}.
\newblock In \emph{2020 28th Signal Processing and Communications Applications
  Conference (SIU)}, pages 1--4.

\end{thebibliography}
